\documentclass[preprint,12pt]{elsarticle}

\usepackage{amssymb}
\usepackage{amsmath}
\usepackage{algorithm,algorithmicx,algpseudocode}
\usepackage{booktabs}      
\usepackage{multirow}      
\usepackage{tabularx}      
\usepackage{adjustbox}
\usepackage{wrapfig}
\usepackage{color,soul}

\begin{document}

\begin{frontmatter}



\title{Deep Positive-Negative Prototypes for Adversarially Robust Discriminative Prototypical Learning} 


\author[1]{Ramin Zarei Sabzevar}
\ead{rzarei@mail.um.ac.ir}

\author[1]{Hamed Mohammadzadeh}
\ead{hamedmohammadzadeh@mail.um.ac.ir}

\author[1]{Tahmineh Tavakoli}
\ead{ta.tavakoli@mail.um.ac.ir}

\author[1]{Ahad Harati\corref{cor1}}
\ead{a.harati@um.ac.ir}

\cortext[cor1]{Corresponding author}

\affiliation[1]{organization={Computer Engineering Department, Ferdowsi University of Mashhad},
            city={Mashhad}, country={Iran}}

\begin{abstract}
Despite the advantages of discriminative prototype-based methods, their role in adversarial robustness remains underexplored.
Meanwhile, current adversarial training methods predominantly focus on robustness against adversarial attacks without explicitly leveraging geometric structures in the latent space, usually resulting in reduced accuracy on the original clean data. We propose a novel framework named Adversarially trained Deep Positive-Negative Prototypes (Adv-DPNP), which integrates discriminative prototype-based learning with adversarial training.
Adv-DPNP uses unified class prototypes that serve as both classifier weights and robust anchors in the latent space.
Moreover, a novel dual-branch training mechanism maintains stable prototypes by updating them exclusively with clean data, while the feature extractor is trained on both clean and adversarial inputs to increase invariance to adversarial perturbations. 
In addition, we use a composite loss that combines positive-prototype alignment, negative-prototype repulsion, and consistency regularization to further enhance discrimination, adversarial robustness, and clean accuracy. 
Extensive experiments on standard benchmarks (CIFAR-10/100 and SVHN) confirm that Adv-DPNP improves clean accuracy over state-of-the-art defenses and baseline methods, while maintaining competitive or superior robustness under a suite of widely used attacks, including FGSM, PGD, C\&W, and AutoAttack. We also evaluate robustness to common corruptions on CIFAR-10-C, where Adv-DPNP achieves the highest average accuracy across severities and corruption types. 
Additionally, we provide an in-depth analysis of the discriminative quality of the learned feature representations, highlighting  the effectiveness of Adv-DPNP in maintaining compactness and clear separation in the latent space.
\end{abstract}



\end{frontmatter}


\section{Introduction}
\label{sec1}

Prototype-based learning has long been recognized as an interpretable and geometrically principled approach for multiclass classification by representing each class with one or more prototypes \cite{zarei2023prototype,Snell2017,Kohonen1990}. However, these methods historically have been less competitive in natural (clean) accuracy compared to modern deep neural networks (DNNs), which dominate tasks such as image classification \cite{dosovitskiy2020vit,he2016deep}, speech recognition \cite{graves2013speech}, and natural language processing \cite{devlin2019bert}. Recently, discriminative extension of prototype-based learning is proposed which closes the accuracy gap with DNNs, while offering compact and well-separated decision regions \cite{zarei2025deep}. 
However, DNNs remain highly susceptible to carefully crafted adversarial perturbations added to their input, which can lead to a significant drop in accuracy \cite{szegedy2014intriguing}.
This vulnerability is particularly critical in safety-sensitive applications such as autonomous driving \cite{ibrahum2025deep}, medical diagnosis \cite{finlayson2019adversarial}, and security-sensitive systems \cite{li2024comprehensive}, where models must sustain both high accuracy and robustness. 

To mitigate the risks posed by adversarial vulnerabilities, various defense strategies have been developed by training models on adversarially perturbed examples \cite{Goodfellow2015Explaining,madry2018towards,zhang2019theoretically,Wang2020Improving,cui2024decoupled}.
Despite their effectiveness, these approaches primarily focus on improving robustness against adversarial perturbations without explicitly considering the underlying geometric structure of the feature space. Therefore, they might suffer from reduced clean accuracy on the original data and suboptimal generalization due to inadequate intra-class compactness and inter-class separation.
In such Adversarial Training (AT) approaches, the gradients from perturbed samples may detrimentally shift decision boundaries, leading to reduced stability and discrimination.

Accordingly, some recent works have explored using prototypes and geometric constraints to address these issues \cite{xu2022orthogonal,fang2024towards,mygdalis2022hyperspherical,mustafa2021deeply}. 
Nevertheless, these approaches either fix classifier weights, which limits the model’s ability to adapt margins to data complexity, or keep classifier weights and prototypes as separate parameter sets, leading to misalignment and unstable decision boundaries \cite{Kansizoglou2023}. Another issue is updating prototypes using adversarial samples, which contaminates the prototypes with adversarial noise and reduces their stability.

Motivated by these observations, we propose a novel framework named Adversarially trained Deep Positive-Negative Prototypes (Adv-DPNP), which extends our discriminative prototype-based learning paradigm \cite{zarei2025deep} to the adversarial domain. Adv‑DPNP is designed to enforce adversarial robustness into the learning process while preserving the concept of class prototypes and maintaining a well-structured latent space. In our approach, each class is represented by a positive prototype that serves as a robust anchor and acts as classifier weights for the corresponding class at the same time, while the nearest rival class centers are treated as negative prototypes. Hence, a compact and well-separated feature space is formed, which combines the benefits of prototype-based methods and discriminative learning.

Additionally, we propose a novel dual-branch training mechanism that leverages both clean and adversarial data through two separate branches: the clean-data branch processes original samples and updates both class prototypes and the feature extractor parameters, while the adversarial-data branch handles adversarially perturbed data and only updates the feature extractor parameters and keeps prototypes unchanged. This dual-branch approach ensures that the learned prototypes capture the distribution of clean data, while the feature extractor learns robust representations by encouraging the alignment of corresponding clean and adversarial inputs in the latent space.

The main contributions of this paper are summarized as follows:
\begin{itemize}
\item{We propose Adv-DPNP, a novel framework that integrates discriminative prototype-based learning with adversarial training. Adv-DPNP explicitly unifies classifier weights and prototypes into shared parameters, ensuring that decision boundaries and geometric class centers remain inherently aligned in adversarial settings.}
\item{We introduce a novel dual-branch training mechanism to exclusively update the prototypes using only the original clean data, while training the feature extractor on clean and adversarial data together to obtain robust feature representations. This dual update scheme ensures prototype stability and prevents adversarial perturbations from affecting the prototypes.}
\item{We develop a composite loss function that unifies cross-entropy with positive prototype alignment, negative prototype repulsion, and consistency regularization terms. This loss effectively balances clean-data accuracy and adversarial robustness, establishing a structured latent space robust to perturbations.}
\end{itemize}

Through extensive experiments conducted on widely used benchmarks, CIFAR-10/100 \cite{krizhevsky2009learning} and SVHN \cite{netzer2011reading}, we demonstrate that Adv‑DPNP consistently outperforms state-of-the-art (SOTA) adversarial training methods in terms of clean accuracy and robustness. Also, we provide an in-depth analysis of the discriminative quality of learned feature representations using several quantitative metrics, highlighting the effectiveness of Adv-DPNP in maintaining compactness and clear separation in latent space.
The rest of this paper is organized as follows: Section \ref{sec2} reviews related work on adversarial attacks, adversarial training defenses, and discriminative prototype-based learning. In Section \ref{sec3}, we present the Adv-DPNP framework, detailing our dual-branch training strategy, unified prototype design, and composite loss functions. Section \ref{sec4} provides experimental results, comparisons with SOTA methods, and a detailed analysis of feature space discriminability. Finally, Section \ref{sec5} concludes the paper and discusses directions for future research. 

\section{Related Work}\label{sec2}
The field of adversarial robustness has been extensively studied, leading to the development of numerous attack and defense mechanisms \cite{wang2023adversarial}. This section provides an overview of related works, categorizing them into three key areas. First, we review the evolution of adversarial attacks, designed to challenge DNNs. Next, we discuss major defense strategies, including adversarial training and various regularization and feature-manipulation techniques. Finally, we explore discriminative prototype-based approaches that aim to enhance both interpretability and robustness by leveraging structured feature representations. 
These insights collectively establish the foundation for our proposed method, which integrates adversarial training with discriminative prototype-based learning to improve clean accuracy while maintaining robustness against adversarial perturbations and common corruptions.

\subsection{Adversarial Attacks} 
Adversarial attacks have emerged as one of the most critical challenges in deep learning research. Early work by Szegedy et al. \cite{szegedy2014intriguing} revealed that DNNs are vulnerable to perturbations in input data that can drastically change model predictions,  even though such perturbations are imperceptible to human observers.
Goodfellow et al. \cite{Goodfellow2015Explaining} explained that the high-dimensional nature and nearly linear behavior of DNNs are key factors in their susceptibility to adversarial perturbations. Consequently, they proposed the Fast Gradient Sign Method (FGSM) as a computationally efficient approach to generate adversarial examples, which updates input data in the gradient direction of maximizing the loss function, scaled by a small constant to ensure that the perturbed input is imperceptible. However, single-step methods like FGSM tend to underestimate the worst-case adversarial perturbations needed to fool robustly trained models. 

To address this limitation, the Basic Iterative Method \cite{kurakin2017adversarial} and the Projected Gradient Descent (PGD) \cite{madry2018towards} extend FGSM by iteratively applying smaller perturbation steps.
PGD optimizes adversarial perturbations within a $l_p$-bounded region and is widely considered one of the strongest first-order attacks. By integrating a momentum term into the iterative process \cite{dong2018boosting}, these attacks can escape local minima more effectively.

Other notable methods include DeepFool \cite{Moosavi2016DeepFool}, which seeks the minimal perturbation needed to cross the decision boundaries, and the Carlini and Wagner (C\&W) attack \cite{carlini2017towards}, which formulates adversarial perturbation generation as an optimization problem with explicit distance constraints.
More recently, Croce and Hein \cite{croce2020reliable} introduced AutoAttack (AA), an ensemble of parameter-free attacks including PGD variants, a targeted decision-boundary attack, and a black-box attack. Its design ensures that no single type of attack is overlooked, making it a robust and standardized tool for benchmarking adversarial defenses. 

Despite extensive research on adversarial attacks, effectively defending against these threats remains challenging, especially when considering unknown perturbation scenarios or attacks under different norm constraints. These challenges underline the necessity for further research into defense mechanisms capable of mitigating diverse adversarial threats.

\subsection{Defense Strategies}
Following the significant advances in adversarial attack methods, another line of research has focused on defense strategies.
One of the most influential defense approaches is AT \cite{Goodfellow2015Explaining,madry2018towards}, in which models are explicitly trained on perturbed inputs. Such methods adopt a min-max optimization procedure, where the inner maximization generates worst-case perturbations that maximize the model’s loss, while the outer minimization updates the model parameters to be resilient against these perturbations.

Subsequent methods, such as AT-HE \cite{pang2020boosting}, integrate AT with hypersphere embedding by normalizing features in the penultimate layer and weights in the softmax layer with an additive angular margin.
Additional methods, such as Adversarial Logit Pairing \cite{kannan2018adversarial} and feature scattering techniques \cite{zhang2019defense}, aim to enforce similarity between feature representations of clean and adversarial samples.
Moreover, MART \cite{Wang2020Improving} specifically treats misclassified adversarial samples by minimizing the distributional discrepancy between them and their clean counterparts, thereby focusing training on hard examples.
Furthermore, TRADES \cite{zhang2019theoretically} extends AT by adding a Kullback–Leibler (KL) regularization term that minimizes the divergence between the predictions on clean and adversarial examples, thus encouraging the model to balance the trade-off between clean accuracy and adversarial robustness.
Building on the same idea, the Improved KL (IKL) loss \cite{cui2024decoupled} reformulated the asymmetric KL objective as the sum of a weighted mean-squared error and a soft-label cross-entropy; hence eliminating the optimization asymmetry leading to better class-wise global statistics and mitigation of sample-wise bias.  

A separate line of work argues that the weak accuracy–robustness trade-off arises because vanilla AT excessively increases the decision-boundary margin.
Helper-based Adversarial Training (HAT) \cite{rade2022hat} addresses this by adding helper samples (extrapolated adversarial images given soft labels from a standard model) to pull the boundary back toward informative directions, boosting clean accuracy with negligible robustness loss.  
Follow-up schemes pursue the same intuition, Increasing-Margin Adversarial Training \cite{ma2023ima} adaptively enlarges or shrinks margins through an equilibrium analysis and has been adopted in medical-imaging pipelines. 
Analytic tools such as ATVis \cite{zhu2024atvis} later visualized these margin dynamics and confirmed HAT’s findings.  
Our Adv-DPNP regularizes the margins via prototypes and discriminative terms in its loss function and therefore remains compatible with these works.

Beyond loss-level regularization, several researchers attempt to suppress or adapt internal feature activations corrupted by adversarial noise. Feature Denoising \cite{xie2019feature}
is proposed to denoise intermediate maps via non-local means and bilateral filtering, while Channel Activation Suppression \cite{bai2021cas} zeros out non-robust channels during training. CIFS \cite{yan2021cifs} goes further by selecting channels according to a learned importance mask that suppresses less-relevant features.
Most recently, Kim \emph{et al.} introduced Feature Separation and Recalibration (FSR) \cite{kim2023fsr}, which first separates robust and non-robust activations via a Gumbel-Softmax learnable mask and then recalibrates the non-robust branch instead of discarding it.  
In contrast to these activation-level methods, our Adv-DPNP keeps all activations but organizes the latent space through prototype anchors; thereby, it can be complemented with such feature-manipulation modules without extra architectural overhead. 

Despite these efforts, existing adversarial training approaches focus primarily on reducing adversarial error rates, paying less attention to how the geometric structures of the feature space are formed. Consequently, while these models can exhibit robust performance against adversarially perturbed data, they may lack optimal intra-class compactness and inter-class separation, potentially harming their performance on clean data and limiting interpretability. These observations motivate the exploration of alternative solutions, particularly those integrating adversarial robustness with discriminative prototype-based learning to capture robust structured geometric representations.

\subsection{Prototype-Based learning and Discriminative Approaches}
Prototype-based learning methods have gained significant attention due to their ability to provide clear and interpretable representations of class structures \cite{zarei2023prototype}.
These methods represent each class with one or more prototypes that capture the underlying structure and intrinsic geometry of their corresponding data. 
Moreover, prototype-based methods inherently provide robustness against outliers by rejecting ambiguous or out-of-distribution samples and reducing sensitivity to individual anomalous instances, since classification decisions depend primarily on the proximity of samples to learned prototypes \cite{Yang2022Convolutional}.
Despite these interpretability and robustness advantages, traditional prototype-based methods often struggle to achieve classification accuracies comparable to discriminative approaches. In particular, early methods, such as Learning Vector Quantization \cite{Kohonen1990} and Self-Organizing Maps \cite{Kohonen1998}, defined prototypes directly in the original input space, limiting their ability to capture complex or high-level patterns within the data.

More recently, prototype-based representations have been integrated with DNNs to enhance discriminative power, interpretability, and generalization.
For example, Prototypical Networks \cite{Snell2017} compute class prototypes as the mean of feature representations of samples belonging to the same class within the latent space.
Other related methods using discriminative loss functions, such as Center Loss (CL) \cite{Wen2016, Shi2023}, explicitly minimize the distances between feature representations and their corresponding class centers. Although these methods enhance the intra-class compactness of learned feature representations, they maintain separate parameter sets and rely on independent update mechanisms for classifier weights and prototypes, resulting in misalignment between these components and consequently insufficient inter-class separation \cite{zarei2025deep,Kansizoglou2023}.
In response to these limitations, we recently proposed the DPNP framework \cite{zarei2025deep}, a unified representation that merges classifier weights and class prototypes into shared parameters. By unifying classifier weights with class prototypes, decision boundaries naturally align with the learned prototypes, enhancing intra-class compactness and inter-class separability and leading to a well-structured feature space.
Furthermore, this approach eliminates the need to store separate vectors for the class prototypes and classifier weights, thereby preventing potential mismatches between these components \cite{Kansizoglou2023}.

Despite the success of both discriminative prototype-based learning and adversarial training independently, the integration of these two paradigms remains relatively underexplored. Although standard adversarial training methods effectively improve robustness, they typically neglect the geometric structure of the latent space. As a consequence, the learned features may not be optimally structured for clear class separation, which can limit interpretability and potentially reduce performance on clean data. 
To address these issues, several studies have attempted to investigate the role of discriminative loss functions such as CL and triplet loss in improving adversarial robustness \cite{mao2019metric,hosler2021learning}. 
Similarly, methods proposed in \cite{mygdalis2022hyperspherical,mustafa2021deeply} have further expanded this concept by applying geometric constraints, including hyperspherical or convex polytope constraints, on intermediate network layers to enhance intra-class compactness and inter-class separation. However, these methods incorporate adversarial examples directly into the prototype update process and rely on separate parameter sets for network weights and prototypes, which can lead to misalignment between them.
Consequently, these design choices can negatively impact prototype stability \cite{Kansizoglou2023} and ultimately limit their ability to achieve robustness comparable to SOTA defense, particularly under white-box attack scenarios where an adversary has complete access to model gradients.

An alternative line of work tackles this misalignment by fixing or regularizing the classifier weights themselves.  
Orthogonal Classifier (OC) \cite{xu2022orthogonal} sets the softmax layer to a precomputed, fully dense orthogonal matrix with equal magnitude entries, enforcing uniform angular separations for robustness improvement.  
Building on this idea, Diversity via Orthogonality (DIO) \cite{fang2024towards} augments the network with multiple parallel classifier heads, then injects layer-wise orthogonality regularizers together with a margin-maximization loss to promote feature diversity and robustness.
These orthogonality-based schemes enlarge decision margins through fixed geometric constraints rather than learnable prototypes. Unlike OC and DIO, which fix or strictly regularize the classifier weights, our Adv-DPNP keeps the prototypes trainable, allowing flexible margin adjustment while preserving classifier weights-prototypes alignment.

\section{Proposed Method}\label{sec3}
In this section, we present our Adv‑DPNP framework, which seamlessly integrates adversarial training with discriminative prototype-based learning. Our approach utilizes shared parameters for the class prototypes and the classifier weights. These prototypes are learned through a composite loss function and optimized using a novel dual-branch training mechanism that processes clean and adversarial data differently. The strategy ensures that the feature extractor outputs discriminative representations and generates an organized latent space, even when the input is perturbed.

We first present the necessary notations, followed by the detailed formulation of each loss component, and finally explain how separate update rules are implemented for clean and adversarial data.

\subsection{Notations and Problem Formulation}\label{subsec3-1}

Let $\mathcal{D} = \{(x_i, y_i)\}_{i=1}^{N}$ be a labeled dataset consisting of $N$ samples, where each input $x_i \in \mathbb{R}^D$ is associated with a corresponding class label $y_i \in \{1,2,\dots,M\}$ for a classification task with $M$ classes. Our goal is to learn a robust classification model that remains accurate on clean data while resisting adversarial perturbations.
We denote the deep feature extractor by the function $f(x;\theta): \mathbb{R}^D \rightarrow \mathbb{R}^d$ parameterized by $\theta$, which maps each input sample $x$ to its corresponding feature representation in a latent space of dimension $d$. In adversarial settings, an adversarial example $\tilde{x} = x + \delta$ is generated by perturbing each clean input $x$ with a bounded adversarial noise $\delta$ using an adversarial attack such as PGD \cite{madry2018towards}, subject to $\|\delta\| \leq \epsilon$, where $\epsilon$ denotes the maximum allowed perturbation magnitude.

In conventional DNN classifiers, the output logits for each class $j$ are computed as the inner product $w_j^Tf(x;\theta)$, where $w_j \in \mathbb{R}^d$ is the weight vector associated with class $j$. Based on the unified prototype-based design of our recent work \cite{zarei2025deep}, we associate each class $j$ with a learnable parameter vector $c_j \in \mathbb{R}^d$, which simultaneously serves as both the class prototype and the classifier weight. 
This shared parameterization explicitly aligns the decision boundaries defined by the classifier weights with the geometric structure of class representations in the feature space. Consequently, these unified class prototypes act as shared anchors, serving both as classifier weights and as geometric centers for their respective classes in the latent space.

\subsection{Composite Loss Function}\label{subsec3-3}
To achieve both high classification accuracy and adversarial robustness, the Adv-DPNP framework combines three key objectives as follows: 
1) Deep Positive Prototype (DPP) for accurate classification and alignment of feature representations with their corresponding class prototypes, 
2) Deep Negative Prototypes (DNP) for pushing class prototypes away from their nearest rivals to improve the separation between them, 
and 3) Deep Feature Alignment (DFA) for ensuring consistent probability outputs between clean and adversarial inputs.

\noindent
\textbf{DPP Loss.} To effectively shape the geometry of the latent space, our prototypes $c_j$ are constrained to lie on the surface of a hypersphere with radius $\alpha$, playing a dual role as class centers (point representatives in the latent space) and as the weight vector of neuron $j$ in the classifier. This constraint allows them to actively influence the organization of feature representations rather than being passively estimated as extra separate parameters, usually through mean calculations \cite{zarei2025deep}.
In our proposed framework, the probability $p_j(x)$ assigned to class $j$ for input $x$ is computed using a softmax function over the similarity scores between the feature representation $f(x; \theta)$ and the prototype $c_j$ relative to other class prototypes $c_k$:
\begin{equation}
\label{eq_prob}
p_{j}(x) = \frac{\exp\left(\frac{c_j^\top f(x; \theta)}{\alpha}\right)}{\sum_{k=1}^{M} \exp\left(\frac{c_k^\top f(x; \theta)}{\alpha}\right)},
\end{equation}
We adopt the cross-entropy (CE) loss as the primary classification objective, encouraging the model to assign high probability to the correct class label $y_i$ for each input sample $x_i$:
\begin{equation}
\label{eq_CE}
L_{\text{CE}}(x_i,y_i) = -\log p_{y_i}(x_i) = -\log \frac{\exp\left(\frac{c_{y_i}^\top f(x_i; \theta)}{\alpha}\right)}{\sum_{k=1}^{M} \exp\left(\frac{c_k^\top f(x_i; \theta)}{\alpha}\right)}
\end{equation}
where $p_{y_i}(x_i)$ denotes the predicted probability assigned to the true class $y_i$ for the sample $x_i$. 
Building upon the unified prototypes ${c_j}$, we define the DPP loss that integrates two components, the CE loss for accurate classification and a positive prototype alignment term, encouraging the feature representations $f(x_i; \theta)$ to closely align with their corresponding class prototypes $c_{y_i}$:
\begin{equation}
\label{eq_DPP}
L_{\text{DPP}}(x_i,y_i) = 
- \underset{L_{\text{CE}}(x_i,y_i)}{\underbrace{
\log \frac{\exp\left(\frac{c_{y_i}^\top f(x_i; \theta)}{\alpha}\right)}
{\sum_{k=1}^{M} \exp\left(\frac{c_k^\top f(x_i; \theta)}{\alpha}\right)}
}} + \frac{\lambda_{\text{DPP}}}{2}\|f(x_i; \theta) - c_{y_i}\|_2^2
\end{equation}
where $c_{y_i}$ is the true class prototype and $\lambda_{\text{DPP}}$ controls the trade-off between correct classification and clustering around each class prototype $c_{y_i}$.
By minimizing the Euclidean distance between $f(x_i;\theta)$ and the corresponding positive prototype $c_{y_i}$, the features are pulled closer to the correct class center, promoting intra-class compactness in the latent space. 
In addition, at the start of each epoch, all prototypes $c_j$ are renormalized to keep them on a hypersphere of radius $\alpha$ (i.e., $\|c_j\|_2=\alpha$); hence, $1/\alpha$ acts as a temperature scaling factor that helps to obtain well-behaved CE gradients and improve numerical stability.

\noindent
\textbf{DNP Loss.} In addition, we incorporate the DNP loss to maximize inter-class margins by pushing prototypes away from each other. In this paper, we consider only the nearest rival prototype as the negative prototype for the corresponding class. Formally, $c_{\text{j}}^{\text{neg}}$ for each positive prototype $c_j$ is defined as:
\begin{equation}
\label{eq_C_Neg}
c_j^{\text{neg}} \leftarrow c_{\underset{k \ne j}{\operatorname*{arg\,min}} \left\| c_k - c_j \right\|}
\end{equation}
Hence, the DNP loss is defined as:
\begin{equation}
\label{eq_DNP}
L_{\text{DNP}} = -\frac{1}{M} \sum_{j=1}^{M} \|c_j - c_{j}^{\text{neg}}\|_{1/2}^{1/2}
\end{equation}
This term is sample-independent and is computed individually for each class. Hence, it is applied once per mini-batch, according to the positive prototype update rate. We select the norm $\ell_{1/2}$ to emphasize short-range repulsion between nearest prototypes. This exerts stronger repulsion when the centers are relatively close, effectively increasing the separations among the nearest classes.

\noindent
\textbf{DFA Loss.} While DPP and DNP collectively improve classification and impose geometric constraints on prototypes, they do not explicitly ensure consistency between clean and adversarial representations derived from feature extractor output. We incorporate the DFA loss $L_{\text{DFA}}(x_i, \tilde{x_i})$ derived from the Kullback-Leibler divergence, similar to TRADES \cite{zhang2019theoretically}, which penalizes discrepancies in predicted distributions for clean inputs $x_i$ and their adversarial counterparts $\tilde{x_i}$:
\begin{equation}
\label{eq_DFA}
L_{\text{DFA}}(x_i, \tilde{x_i}) = \sum_{j=1}^{M} p_{j}(x_i) \log \frac{p_{j}(x_i)}{p_{j}(\tilde{x_i})}.
\end{equation}
Minimizing $L_{\text{DFA}}$ encourages the model to generate similar probabilistic predictions for clean and adversarial input pairs, helping to reduce performance drops during attacks.

\noindent
\textbf{Adv-DPNP Loss.} By combining these components, the overall loss function of our Adv‑DPNP framework is formulated as a weighted sum:
\begin{equation}
\label{eq_Adv_DPNP}
L_{\text{Adv‑DPNP}} = \lambda_{\text{DNP}}\,L_{\text{DNP}} + \sum_{i=1}^{N}{\frac{L_{\text{DPP}}(x_i,y_i) + L_{\text{DPP}}(\tilde{x_i},y_i) + \lambda_{\text{DFA}}\,L_{\text{DFA}}(x_i, \tilde{x_i})}{2N}}
\end{equation}
where $\lambda_{\text{DNP}}$ and $\lambda_{\text{DFA}}$ are hyperparameters determining the relative contributions of DNP and DFA losses. 
This composite formulation implements a prototype-based learning framework which generalizes AT/TRADES losses. 
Here DPP serves as the classification objective and explicitly induces intra-class compactness; DNP acts as a margin regularizer at the prototype level, whereas DFA enforces clean–adversarial distributional consistency without architectural overhead. The result is a geometrically well-structured latent space.

It should be noted that in our framework, although adversarial examples $\tilde{x_i}$ pass through the same set of loss terms as clean inputs $x_i$, the prototypes $c_j$ only receive gradient updates from the clean data. This design choice stabilizes the prototype geometry by preventing adversarial perturbations from shifting class centers. In the next section (Section~\ref{subsec3-2}), we elaborate on this dual-branch training procedure and explain how updates are decoupled for clean and adversarial data.

\subsection{Dual-Branch Training Procedure}\label{subsec3-2}
\begin{figure}[t]
\centering
\includegraphics[width=5.3in]{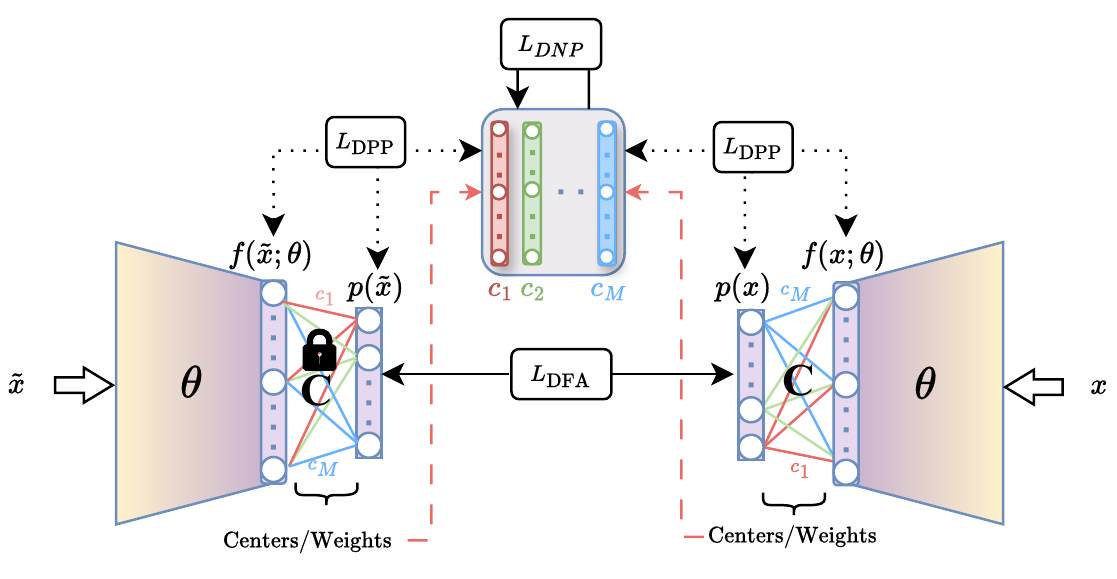}
\caption{Schematic overview of the Adv-DPNP framework illustrating its dual-branch training procedure. The left branch processes adversarial examples $\tilde{x_i}$, where prototypes $(c_1,c_2,\dots,c_M)$ are locked (no gradient flow to prototypes), ensuring adversarial perturbations do not shift the class centers. The right branch handles clean inputs $x$, updating both the feature extractor $\theta$ and the prototypes. Three losses including DPP (promoting intra-class compactness), DNP (pushing prototypes apart for inter-class separation), and DFA (aligning outputs of clean and adversarial data) are jointly optimized to achieve robust and discriminative representations.}
\label{fig_dual_branch}
\end{figure}

While the prototypes $c_j$ serve as universal anchors for both clean and adversarial samples, our training strategy employs a dual-branch scheme that treats these two types of data differently, see Figure \ref{fig_dual_branch}. This approach ensures that the learned model accurately captures the true class prototypes from the clean data while also achieving robustness against adversarial perturbations. 

\noindent
\textbf{Clean Data Branch.} In this branch designated for clean inputs $x_i$, both the feature extractor parameters $\theta$ and the corresponding prototypes $c_{y_i}$ are updated during backpropagation. So, this branch adapts the prototypes to clean inputs, ensuring well-structured class clusters in the feature space and maintaining stable representations of each class center.

\noindent
\textbf{Adversarial Data Branch.} In this branch, we process adversarial examples $\tilde{x_i}$, generated from clean samples $x_i$ using an attack method such as PGD. Here, the prototypes $c_j$ are locked, and the gradient flow from these adversarial inputs to prototypes $c_j$ is blocked; hence they remain fixed in this branch. Therefore, only the feature extractor parameters $\theta$ are updated so that adversarial features align better with the corresponding prototype $c_{y_i}$. This design prevents adversarial perturbations from misplacing the class centers, thus retaining a stable geometric layout derived from clean samples.

By decoupling updates of the prototypes and the feature extractor, Adv-DPNP enforces a stable prototype representation that guides the feature extractor to learn robust features from both clean and adversarial distributions, ensuring consistent alignment with the true class prototypes.
Consequently, the model learns to classify perturbed samples effectively without distorting the established prototype geometry. By combining the steps described in our dual-branch training procedure with the proposed loss functions, Adv-DPNP learns a robust and discriminative representation for both clean and adversarial data. 
\textbf{Algorithm~\ref{algo1}} summarizes the complete training process.

\begin{algorithm}
\small
\caption{Training Algorithm for Adv-DPNP Model}\label{algo1}
\begin{algorithmic}[1]
\Require Training dataset $\mathcal{D}=\{(x_i, y_i)\}_{i=1}^N$, number of epochs $E$, learning rates $\eta$, batch size $B$, perturbation bound $\epsilon$, hyperparameters $\lambda_{\text{DPP}}$, $\lambda_{\text{DFA}}$, $\lambda_{\text{DNP}}$, radius $\alpha$.
\State Initialize network parameters $\theta$ and class prototypes $\{c_j\}_{j=1}^M$
\For {$\text{epoch} = 1$ to $E$}
    \For{each class $j = 1, \dots, M$}
        \State Normalize class prototypes: $c_j \leftarrow \alpha \frac{c_j}{\| c_j \|}$
    \EndFor
    \For{each mini-batch $\mathcal{B}=\{(x_i, y_i)\}_{i=1}^B \subset \mathcal{D}$}
        \For{each sample $(x_i, y_i)$ in the mini-batch $\mathcal{B}$}
            \State Compute feature representation $f(x_i; \theta)$ and probabilities $p_j(x_i)$  
            \State Generate adversarial examples $\tilde{x_i}$            
            \State Compute feature representations $f(\tilde{x_i};\theta)$ and probabilities $p_j(\tilde{x_i})$
            \State Compute loss functions $L_{\text{DPP}}({x_i},{y_i})$, $L_{\text{DPP}}(\tilde{x_i},{y_i})$ and $L_{\text{DFA}}({x_i},\tilde{x_i})$
        \EndFor
        \For{each class $j = 1, \dots, M$}
          \State Find class negative prototype: $c_j^{\text{neg}}$ using \eqref{eq_C_Neg}
        \EndFor
        \State Compute the loss function $L_{\text{DNP}}$ based on \eqref{eq_DNP}
        \State Compute the loss function $L_{\text{Adv-DPNP}}$ based on \eqref{eq_Adv_DPNP}
        \State Update feature extractor parameters:
         $\theta \leftarrow \theta - \eta \nabla_\theta L_{\text{Adv-DPNP}}$         
        \For{each class $j = 1, \dots, M$}
        \State Update unified class $j$ prototype:    
        \State $c_j \leftarrow c_j - \eta \nabla_{c_j}(\lambda_{\text{DNP}}L_{\text{DNP}}+\frac{1}{2B}\sum_{i}^{B}L_{\text{DPP}}(x_i, y_i)+\lambda_{\text{DFA}}L_\text{DFA}(x_i, \tilde{x_i}))$
        \EndFor
    \EndFor
\EndFor
\end{algorithmic}
\end{algorithm}

\section{Experiments}\label{sec4}
In this section, we extensively evaluate Adv-DPNP on three widely used datasets, namely CIFAR-10 \cite{krizhevsky2009learning}, CIFAR-100, and SVHN \cite{netzer2011reading}. For a comprehensive comparison, we conduct experiments using two network backbones: ResNet-18 (RN-18) and WideResNet-34-10 (WRN-34-10).
Section~\ref{subsec41} explains our experimental setup, including parameter settings and training protocols. In Section~\ref{subsec42}, we present a comparative robustness analysis of Adv-DPNP against several prominent baseline methods, including AT \cite{madry2018towards}, TRADES \cite{zhang2019theoretically}, MART \cite{Wang2020Improving}, and AT\_HE \cite{pang2020boosting}, as well as SOTA methods such as OC \cite{xu2022orthogonal}, PCL \cite{mustafa2021deeply}, FSR \cite{kim2023fsr}, HAT \cite{rade2022hat}, DIO \cite{fang2024towards}, and IKL \cite{cui2024decoupled}.
Section~\ref{subsec43} analyzes the effectiveness of Adv-DPNP in structuring the learned feature space for discriminative representations. Section~\ref{subsec44} examines the robustness under common image corruptions. Finally, Section~\ref{subsec45} presents experiments designed to rule out the possibility of gradient obfuscation in Adv-DPNP.

\subsection{Experimental Setup}\label{subsec41}
\noindent \textbf{Training Parameters.}
We follow the recommended settings from \cite{pang2021bag} for adversarially training the models to prevent the training parameters from undermining robustness improvements. Specifically, all evaluated models are trained with a weight decay of $5 \times 10^{-4}$ for 110 epochs. We use stochastic gradient descent with a momentum of 0.9 and an initial learning rate of 0.1, which is reduced to 0.01 at epoch 100 and 0.001 at epoch 105. The batch size is set to 128 in all experiments. We also apply standard data augmentation during training, namely random cropping and horizontal flipping. All models, including Adv-DPNP and all baseline/SOTA defenses, are trained from scratch under this training protocol, using the authors’ official implementations to ensure a fair comparison.

\noindent \textbf{Adversarial Training Setup.}
For adversarially training the models, we follow standard practice and generate adversarial examples using the PGD attack \cite{madry2018towards} under the $l_{\infty}$ norm bound $\epsilon = 8/255$ for 10 iterations. The step size is set to $2/255$ for CIFAR-10/100 and $1/255$ for SVHN. This setup is used for all adversarially trained methods to ensure comparability.

\noindent
\textbf{Hyperparameters.}
For all evaluated methods, we use the hyperparameters suggested in their respective papers. 
For Adv-DPNP, we set the hyperspherical radius to $\alpha=40$ for all experiments. 
On CIFAR-10 and SVHN, we use $\lambda_{\text{DPP}}=0.1$, $\lambda_{\text{DNP}}=0.1$, and $\lambda_{\text{DFA}}=2$, while on CIFAR-100, we set $\lambda_{\text{DPP}}=0.3$, $\lambda_{\text{DNP}}=3$, and $\lambda_{\text{DFA}}=2.4$ to account for its larger number of classes and tighter class separations. These values are kept constant across all runs and comparisons. 

\subsection{Robustness Against Adversarial Perturbations}\label{subsec42}

\noindent\textbf{Evaluation Details.}
We evaluate the adversarial robustness of Adv-DPNP using a variety of widely adopted attacks. To thoroughly assess the robustness from multiple perspectives, we perform comparisons across two complementary sets of rival methods.
The first set, which we refer to as \emph{baselines}, consists of famous defense methods which are tested under the standard PGD attacks constrained by $l_{\infty}$ ($\epsilon=8/255$, with a step size of $1/255$ for 20 iterations), $l_2$ ($\epsilon=128/255$ with a step size of $16/255$ for 20 iterations), and $l_1$ ($\epsilon=2000/255$ with a step size of 50 for 100 iterations) norms, along with AutoAttack (AA), which is an ensemble of three white-box attacks and one black-box attack. 
We further evaluate our Adv-DPNP against the second set which includes recently published SOTA approaches under widely used diverse adversarial attacks, including FGSM, PGD (20 and 100 iterations), C\&W (30 iterations with a confidence margin of 50), and an Ensemble attack.
The Ensemble accuracy is calculated by sequentially applying FGSM, PGD-20, PGD-100, and C\&W attacks to each test sample, marking it as failed if its predicted label changes under any of these attacks.

\begin{table}[!t]
\setlength{\tabcolsep}{3pt}      
\renewcommand{\arraystretch}{1.15} 
\scriptsize 
\begin{tabular*}{\textwidth}{@{\extracolsep\fill}lccc|ccc|ccc|ccc}
\toprule
& \multicolumn{6}{c|}{\textbf{CIFAR‑10}} 
& \multicolumn{6}{c}{\textbf{CIFAR‑100}} \\
\cmidrule(lr){2-7}\cmidrule(lr){8-13}
\textbf{Method} 
& Clean & AA & Avg & \multicolumn{3}{c|}{PGD} 
& Clean & AA & Avg & \multicolumn{3}{c}{PGD} \\          
\cmidrule(lr){2-2}\cmidrule(lr){3-3}\cmidrule(lr){4-4}\cmidrule(lr){5-7}
\cmidrule(lr){8-8}\cmidrule(lr){9-9}\cmidrule(lr){10-10}\cmidrule(lr){11-13}
& – & $l_\infty$ & – & $l_1$ & $l_2$ & $l_\infty$  
& – & $l_\infty$ & – & $l_1$ & $l_2$ & $l_\infty$ \\     
\midrule
ST        & \textbf{94.39} &  0.00 & 47.20 & 40.77 &  0.04 &  0.00 
          & \textbf{76.24} &  0.00 & 38.12 & 25.90 &  1.42 &  0.03  \\
AT        & 83.78 & 48.20 & \underline{65.99} & 51.04 & \underline{63.72} & 52.75 
          & 57.91 & 24.98 & 41.45 & 38.03 & \underline{38.95} & 29.17  \\ 
AT\_HE    & 80.28 & 47.29 & 63.79 & 48.44 & 60.67 & 50.77 
          & 57.11 & \textbf{25.73} & 41.42 & 36.71 & 37.99 & 29.35  \\
TRADES    & 81.38 & \underline{49.00} & 65.19 & 49.72 & 62.37 & \underline{53.18} 
          & 57.99 & 25.06 & \underline{41.53} & \underline{38.43} & 38.77 & \underline{30.17}  \\
MART      & 79.60 & 48.18 & 63.89 & \underline{52.39} & 61.41 & 51.94 
          & 53.64 & \underline{25.61} & 39.63 & 34.38 & 37.66 & 29.46  \\
\textbf{Adv‑DPNP}  & \underline{84.63} & \textbf{49.18} & \textbf{66.91} & \textbf{55.11} & \textbf{64.13} & \textbf{53.68} 
          & \underline{59.77} & 25.39 & \textbf{42.58} & \textbf{39.26} & \textbf{40.02} & \textbf{30.80}  \\
\bottomrule
\end{tabular*}
\caption{Comparison of clean and adversarial accuracy (\%) on CIFAR‑10 and CIFAR‑100 under AutoAttack (AA) and various PGD attacks ($l_1$,$l_2$,$l_{\infty}$). The “Avg” column denotes the average of clean and AA accuracy. Adv-DPNP offers a better balance between clean accuracy and robustness compared to other baseline methods.}
\label{tab:baselines_comp}
\end{table}

\noindent\textbf{Comparison with Baselines.}
In Table \ref{tab:baselines_comp}, we compare the proposed Adv-DPNP with the baseline adversarial defenses on CIFAR-10 and CIFAR-100. For reference, the standard training (ST) model (i.e., an undefended baseline) achieves the highest clean accuracy but collapses under attacks (AA $\approx$ 0\%); thus, we compare clean accuracy only among adversarial defenses.
As shown in Table \ref{tab:baselines_comp}, Adv-DPNP increases clean accuracy on both datasets, improving by 0.85\% on CIFAR-10 and 1.78\% on CIFAR-100 compared with other defenses. This improvement suggests that the dual-branch training scheme in Adv-DPNP effectively captures the true class prototypes from clean data. 
Furthermore, Adv-DPNP outperforms the baselines under unforeseen perturbation types on both datasets. In particular, it improves robustness against $l_1$ perturbations by 2.7\% on CIFAR-10 and against $l_2$ perturbations by 1.07\% on CIFAR-100, indicating better transferability of robustness from $l_\infty$ to $l_1$ and $l_2$ norms.
Lastly, considering the average of clean and AA accuracy (denoted by “Avg” in Table \ref{tab:baselines_comp}), Adv-DPNP achieves a better balance between clean accuracy and robustness. Specifically, it attains the highest AA accuracy on CIFAR-10 and improves Avg by 0.92\%, while on CIFAR-100 it increases Avg by 1.05\% and maintains competitive AA performance. 

\begin{table}[!t]
\centering
\resizebox{\textwidth}{!}{%
\begin{tabular}{llccc|ccccc}
\toprule
\multirow{2}{*}{Dataset / Backbone} & \multirow{2}{*}{Method}
  & Clean & AA & Avg & FGSM & PGD-20 & PGD-100 & C\&W & Ensemble \\
\cmidrule(lr){3-10}
& & \multicolumn{8}{c}{Accuracy (\%)}\\
\midrule
\multirow{7}{*}{\textbf{CIFAR-10} / RN-18}
 & OC   & 81.80 & 46.46 & 64.13 & 57.14 & 51.40 & 49.99 & 50.28 & 49.59 \\
 & PCL  & 83.70 & 40.85 & 62.28 & 51.23 & 44.63 & 43.21 & 43.69 & 41.90  \\
 & HAT  & \underline{84.28} & 47.51 & \underline{65.90} & 57.85 & 52.01 & 50.80 & 50.21 & 48.68 \\
 & FSR  & 81.83 & 47.33 & 64.58 & 57.03 & 52.52 & 51.28 & 49.93 & 48.38 \\
 & DIO  & 83.18 & 48.21 & 65.70 & \underline{58.06} & 52.59 & 51.25 & \underline{50.90} & 49.25 \\
 & IKL  & 77.90 & \underline{48.88} & 63.39 & 57.07 & \textbf{54.14} & \textbf{53.65} & 49.93 & \underline{49.61} \\
 & \textbf{Adv-DPNP} & \textbf{84.63} & \textbf{49.18} & \textbf{66.91} & \textbf{58.12} & \underline{53.68} & \underline{52.61} & \textbf{50.97} & \textbf{50.14} \\
\midrule
\multirow{7}{*}{\textbf{CIFAR-100} / RN-18}
 & OC   & 59.24 & 24.49 & \underline{41.87} & \underline{33.02} & 27.36 & 26.10 & 26.43 & 25.81 \\
 & PCL  & 59.08 & 20.61 & 39.85 & 30.64 & 26.75 & 26.07 & 23.77 & 22.18  \\
 & HAT  & \underline{59.60} & 22.79 & 41.20 & 29.61 & 26.35 & 25.44 & 25.05 & 23.81 \\
 & FSR  & 58.04 & 23.38 & 40.71 & 30.64 & 27.24 & 26.40 & 25.72 & 24.19 \\
 & DIO  & 54.39 & 24.50 & 39.45 & 31.83 & 29.35 & 28.79 & \underline{26.63} & 25.68 \\
 & IKL  & 56.62 & \textbf{25.43} & 41.03 & 32.84 & \textbf{31.17} & \textbf{30.70} & 26.42 & \underline{26.29} \\
 & \textbf{Adv-DPNP} & \textbf{59.77} & \underline{25.39} & \textbf{42.58} & \textbf{33.19} & \underline{30.80} & \underline{30.15} & \textbf{27.29} & \textbf{26.58} \\
\midrule
\multirow{4}{*}{\textbf{SVHN} / RN-18}
 & FSR & 92.96 & 45.22 & 69.09 & 65.13 & 52.82 & 49.43 & 48.97 & 47.05  \\
 & DIO & \textbf{93.52} & 45.44 & \underline{69.48} & \underline{66.17} & 52.97 & 49.48 & 49.72 & 47.57  \\
 & IKL & 85.05 & \underline{47.53} & 66.29 & 61.46 & \underline{55.50} & \underline{54.46} & \underline{50.73} & \underline{50.36}  \\
 & \textbf{Adv-DPNP} & \underline{93.06} & \textbf{48.04} & \textbf{70.55} & \textbf{68.43} & \textbf{57.45} & \textbf{54.85} & \textbf{51.54} & \textbf{50.53}  \\
\midrule
\multirow{4}{*}{\textbf{CIFAR-10} / WRN-34-10}
 & FSR  & \underline{86.39} & 51.14 & \underline{68.77} & \underline{61.69} & 55.76 & 54.28 & 54.19 & 52.37 \\
 & DIO  & 85.28 & 51.95 & 68.62 & 60.68 & 54.75 & 53.41 & \underline{54.79} & 52.70 \\
 & IKL  & 82.23 & \textbf{53.10} & 67.67 & 61.50 & \textbf{57.73} & \textbf{57.26} & 54.36 & \textbf{53.87} \\
 & \textbf{Adv-DPNP} & \textbf{87.47} & \underline{52.77} & \textbf{70.12} & \textbf{62.14} & \underline{57.26} & \underline{55.07} & \textbf{54.96} & \underline{53.74} \\
\bottomrule
\end{tabular}}
\caption{Robustness comparison on three datasets and two backbones. Columns list accuracy (\%) on clean data and six diverse attacks. The Ensemble column counts a sample as wrong if any of FGSM, PGD-20, PGD-100 or C\&W succeeds, providing a fast worst-case estimate.  
“Avg” is the mean of Clean and AutoAttack (AA) accuracies. Adv-DPNP attains the highest “Avg” in all four settings and never drops below second place in AA, demonstrating consistent gains over the competing methods.}
\label{tab:robust_avg2}
\end{table}

\noindent \textbf{Comparison with SOTA.} Table \ref{tab:robust_avg2} further compares Adv-DPNP against the SOTA defenses across three datasets (CIFAR-10, CIFAR-100, SVHN) and two backbone architectures (ResNet-18 and WRN-34-10). Several important observations emerge from this evaluation.
\textbf{First}, Adv-DPNP consistently achieves the highest Avg in all four evaluation blocks. Specifically, on CIFAR-10 with ResNet-18, it achieves an Avg of 66.91\%, outperforming the second-best margin-based defense (HAT) by 1.01\%. On CIFAR-100 with ResNet-18, it attains an Avg of 42.58\%, surpassing the next-best method by 0.72\%. On SVHN with ResNet-18, Adv-DPNP again improves the Avg to 70.55\%, outperforming the orthogonality-based defense DIO by 1.07\%. Finally, with WRN-34-10 trained on CIFAR-10, it achieves an Avg of 70.12\%, surpassing the feature-masking method (FSR) by 1.35\%.
\textbf{Second}, Adv-DPNP demonstrates competitive robustness under AA, consistently achieving either the highest or second-highest AA accuracy across all four settings. This indicates that the robustness gains generalize across datasets and architectures.
\textbf{Third}, Adv-DPNP consistently ranks among the top defenses against diverse individual attacks (FGSM, PGD-20, PGD-100, C\&W) as well as under the Ensemble metric. Notably, it achieves the highest Ensemble accuracy in three of the four evaluation settings.

\noindent\textbf{Robustness against different perturbation sizes.} To further evaluate the robustness of adversarially trained models, we report their accuracy against $l_{\infty}$ constrained PGD and AA with different perturbation sizes. As shown in Table \ref{tab3}, Adv-DPNP consistently outperforms other methods across all tested perturbation sizes, further confirming its robust improvement.  

\begin{table}[!t]
\centering
\scriptsize 
\begin{tabular*}{\textwidth}{@{\extracolsep\fill}lcccccc}
\toprule%
\multirow{2}{*}{Method} & \multicolumn{2}{@{}c@{}}{$\epsilon=\frac{2}{255}$} & \multicolumn{2}{@{}c@{}}{$\epsilon=\frac{4}{255}$} & \multicolumn{2}{@{}c@{}}{$\epsilon=\frac{6}{255}$} \\\cmidrule{2-3}\cmidrule{4-5}\cmidrule{6-7}%
{} & PGD & AA & PGD & AA & PGD & AA \\
\midrule
{AT}      & {77.90} & {76.76} & {70.67} & {68.26} & {62.09} & {58.66} \\ 
{TRADES}  & {75.58} & {74.84} & {68.80} & {67.17} & {61.49} & {58.28} \\ 
{AT\_HE}  & {74.10} & {73.48} & {66.94} & {65.63} & {59.30} & {56.73} \\ 
{MART}    & {70.48} & {69.88} & {64.78} & {63.48} & {58.19} & {56.39} \\ 
{FSR}  & {75.70} & {74.40} & {68.51} & {65.82} & {60.62} & {56.90} \\
{DIO}  & {77.28} & {76.37} & {70.04} & {68.11} & {61.78} & {58.72}\\
{IKL}  & {72.74} & {71.74} & {66.83} & {64.60} & {60.69} & {57.18} \\
{Adv\_DPNP}& {\textbf{78.07}} & {\textbf{77.25}} & {\textbf{70.71}} & {\textbf{68.76}} & {\textbf{62.13}} & {\textbf{58.85}} \\ 
\bottomrule
\end{tabular*}
\caption{Adversarial accuracy (\%) evaluated via PGD and AutoAttack (AA) under different $l_{\infty}$ perturbation bounds ($\epsilon\in\{2/255,4/255,6/255\}$)  on CIFAR-10. Adv-DPNP maintains higher robustness across all tested $\epsilon$ values.
}\label{tab3}
\end{table}

Collectively, these observations confirm the effectiveness and broad generalizability of the Adv-DPNP framework. By anchoring adversarially perturbed representations to prototypes learned solely from clean data, Adv-DPNP successfully shrinks the gap between accuracy and robustness, exhibiting robust performance across multiple datasets, attacks, and network architectures.

\subsection{Discriminative Aspects of the Feature Space}\label{subsec43}
To investigate how Adv-DPNP preserves discriminative feature representations under adversarial perturbations, we analyze the learned feature spaces of adversarially trained models explained in Section \ref{subsec41}, comparing Adv-DPNP against conventional defenses. Specifically, we use four complementary metrics including Fisher’s Discriminant Ratio (FDR) \cite{fisher1936use}, Angular Fisher Score (AFS) \cite{SphereFace2017}, Separation to Compactness Ratio (SCR) \cite{zarei2025deep}, and Class Center Angular Separation \cite{zarei2025deep} to evaluate feature discrimination in both clean and adversarial scenarios. These metrics collectively capture different aspects of intra-class compactness and inter-class separation, providing a comprehensive perspective on each model’s ability to maintain a robust and well-organized latent space.

\noindent \textbf{FDR.} This criterion measures the ratio of inter-class variance to intra-class variance in Euclidean space. Let $\mathcal{C}_j = \{x_i\in\mathcal{D} \mid y_i = j\}$ be the set of all samples belonging to class $j$, and $N_j = |\mathcal{C}_j|$ is its cardinality. Then, FDR is computed as:
\begin{equation}
\label{eq:FDR}
\mathrm{FDR} =
\sum_{j=1}^{M}
\frac{\mathrm{tr}\Bigl(S_j^w\Bigr)}{\mathrm{tr}\Bigl(S_j^b\Bigr)}
= \sum_{j=1}^M\,
\frac{
  \mathrm{tr}\Bigl(
    \sum_{\,x_i \in \mathcal{C}_j}
    \bigl(f(x_i;\theta) - \mu_j\bigr)
    \bigl(f(x_i;\theta) - \mu_j\bigr)^\top
  \Bigr)}{
  \mathrm{tr}\Bigl(
    N_j\bigl(\mu_j - \mu\bigr)\bigl(\mu_j - \mu\bigr)^\top
  \Bigr)}.
\end{equation}
where $tr(.)$ is the trace operation, $S_j^w$ and $S_j^b$ denote within-class and between-class scatter matrices for class $j$ respectively, $\mu_j = \frac{1}{N_j}\sum_{x_i \in \mathcal{C}_j}\,f(x_i;\theta)$ is the mean feature vector of class $j$, and $\mu = \tfrac{1}{N} \sum_{i=1}^N \,f(x_i;\theta)$ is the global mean feature vector. 
A lower FDR value indicates a more discriminative feature space, with smaller within-class variance and larger between-class variance.

\noindent \textbf{AFS.} In most DNNs trained with the softmax function, classification is based on the angular similarity between the feature vectors and the classifier weights of each class. This motivates the use of AFS to quantitatively assess the feature discrimination in angular space by computing the ratio of angular within-class scatter to angular between-class scatter:
\begin{equation}
\mathrm{AFS} = \frac{\sum_{j=1}^{M}\sum_{x_i\in \mathcal{C}_j} \Bigl(1 - \cos\langle f(x_i;\theta), \mu_j\rangle\Bigr)}{\sum_{j=1}^{M}N_j\,\Bigl(1-\cos\langle \mu_j, \mu\rangle\Bigr)}.
\end{equation}
This cosine similarity measures how closely each feature vector is aligned with its corresponding class center. A lower AFS implies that features are angularly well-separated and compact, reflecting better class discrimination in angular space.

\noindent \textbf{SCR.} Unlike FDR and AFS, which measure global variance, SCR explicitly considers the distance between each class center and its nearest rival. Then, SCR is given by:
\begin{equation}
\mathrm{SCR} = \frac{1}{M} \sum_{j=1}^{M}\,\frac{\min_{k \ne j}\|c_k - c_j\|}{\frac{1}{N_j}\sum_{x_i \in \mathcal{C}_j}\,\|f(x_i;\theta) - c_j\|}\,.
\end{equation}
where $c_j$ denotes the center for class $j$. For competing methods, classifier weights are used as class centers. A higher SCR indicates that each class is better separated from its nearest rival while maintaining intra-class compactness. 

\noindent \textbf{Class Center Angular Separation.} 
To further quantify the separability of class centers, we compute two metrics based on angular separation: mean angular separation (\textit{MeanSep}) and minimum angular separation (\textit{MinSep}). Specifically, MeanSep measures the average angular distance between each class center and its nearest rival class center, while MinSep evaluates the minimum angular distance between any two class centers in the feature space. These metrics are formally defined as follows:
\begin{equation}
\mathrm{MeanSep} = \frac{1}{M} \sum_{j=1}^{M} \min_{k \neq j} \arccos\left( c_j^\top c_k \right) \times \frac{180}{\pi}.
\end{equation}
\begin{equation}
\mathrm{MinSep} = \min_{k \neq j} \arccos\left( c_j^\top c_k \right) \times \frac{180}{\pi}.
\end{equation}
Both metrics are computed in degrees. Larger MeanSep and MinSep indicate better separation between class centers, implying stronger margin-based discrimination.

\begin{table}[!t]
\centering
\resizebox{\textwidth}{!}{%
\begin{tabular}{llcccccccc}
\toprule
\multirow{2}{*}{Dataset} & \multirow{2}{*}{Method} 
  & \multicolumn{2}{c}{FDR $\downarrow$} 
  & \multicolumn{2}{c}{AFS $\downarrow$} 
  & \multicolumn{2}{c}{SCR $\uparrow$} 
  & \multirow{2}{*}{MeanSep $\uparrow$} 
  & \multirow{2}{*}{MinSep $\uparrow$} \\
\cmidrule(lr){3-4} \cmidrule(lr){5-6} \cmidrule(lr){7-8}
& & cln & adv & cln & adv & cln & adv & & \\
\midrule
\multirow{9}{*}{CIFAR-10}
  & ST        & \textbf{0.029} & 0.597 & \textbf{0.200} & 2.850 & 0.577 & 0.252 & 82.24 & \underline{75.52} \\
  & AT        & 0.082 & \underline{0.161} & 0.351 & \underline{0.584} & 0.599 & 0.613 & 73.85 & 64.92 \\
  & AT\_HE    & 0.139 & 0.271 & 0.782 & 1.608 & 0.332 & \underline{1.103} & \underline{83.42} & 73.21 \\
  & TRADES    & 0.156 & 0.254 & 0.968 & 1.560 & 0.796 & 0.777 & 80.31 & 74.05 \\
  & MART      & 0.099 & 0.189 & 0.390 & 0.599 & 0.654 & 0.713 & 70.65 & 58.49 \\
  & FSR       & 0.128 & 0.247 & 0.602 & 1.094 & 0.442 & 0.480 & 76.24 & 68.20 \\
  & IKL       & 0.140 & 0.245 & 0.798 & 1.336 & \underline{0.948} & 0.957 & 77.34 & 70.93 \\
  & \textbf{Adv-DPNP}  & \underline{0.059} & \textbf{0.084} & \underline{0.292} & \textbf{0.435} & \textbf{1.681} & \textbf{1.615} & \textbf{95.32} & \textbf{94.51} \\
\midrule
\multirow{9}{*}{CIFAR-100}
  & ST        & \textbf{0.011} & 0.114 & \textbf{0.389} & 1.479 & 0.190 & 0.066 & 65.95 & 44.96 \\
  & AT        & 0.023 & 0.041 & 1.336 & 2.194 & 0.258 & 0.257 & 69.44 & 53.02 \\
  & AT\_HE    & 0.023 & 0.043 & 1.034 & 1.705 & \underline{1.114} & \underline{1.015} & 73.39 & 60.02 \\
  & TRADES    & 0.032 & 0.049 & 1.815 & 2.779 & 0.267 & 0.263 & 73.85 & 59.52 \\
  & MART      & 0.024 & \underline{0.038} & 0.797 & \underline{1.084} & 0.240 & 0.243 & 67.77 & 50.49 \\
  & FSR       & 0.026 & 0.042 & 0.769 & 1.091 & 0.153 & 0.154 & 70.01 & 54.72 \\
  & IKL       & 0.024 & 0.040 & 1.279 & 2.036 & 0.313 & 0.309 & \underline{75.65} & \underline{63.36} \\
  & \textbf{Adv-DPNP}  & \underline{0.019} & \textbf{0.029} & \underline{0.572} & \textbf{0.765} & \textbf{1.501} & \textbf{1.461} & \textbf{89.90} & \textbf{89.52} \\
\bottomrule
\end{tabular}
}
\caption{Quantitative evaluation of feature-space discrimination across different methods on CIFAR-10 and CIFAR-100. Arrows ($\uparrow$ / $\downarrow$) indicate whether higher or lower values are better. The best result in each column is in bold, and the second-best is underlined.}
\label{tab4}
\end{table}

\noindent\textbf{Results.} 
We evaluate these metrics on both clean and adversarial features, where adversarial examples are generated using the 20-step $l_{\infty}$-PGD attack. The results for CIFAR-10 and CIFAR-100 are reported in Table~\ref{tab4}.
Across all models, adversarial features exhibit higher FDR and AFS values compared to clean features, as adversarial perturbations tend to shift the feature representations towards incorrect class centers \cite{mao2019metric}. As expected, undefended ST achieves the lowest FDR and AFS values for clean features, but it drastically fails to maintain a consistent latent space structure under adversarial perturbations, leading to poor robustness and geometric stability.
In contrast, adversarially trained models preserve better structure when confronted with adversarial examples.
Notably, our Adv-DPNP exhibits the lowest FDR and AFS values among robust models in both clean and adversarial settings, indicating better Euclidean and angular discrimination. 
Additionally, Adv-DPNP maintains the highest SCR and the best angular margins (MeanSep/MinSep reach $95.32^\circ/94.51^\circ$ on CIFAR-10 and $89.90^\circ/89.52^\circ$ on CIFAR-100, substantially better than other methods), increasing its generalization. These results indicate that Adv-DPNP successfully learns a globally well-separated set of anchors and preserves a discriminative latent space, exhibiting high intra-class compactness and inter-class separation in both clean and adversarial settings. Moreover, the smaller change between corresponding values of clean/adversarial columns in the case of Adv-DPNP suggests that our dual-branch training procedure successfully learns a robust feature extractor and discriminant prototypes.

\begin{figure}[t]
\centering
\includegraphics[width=1.0\textwidth]{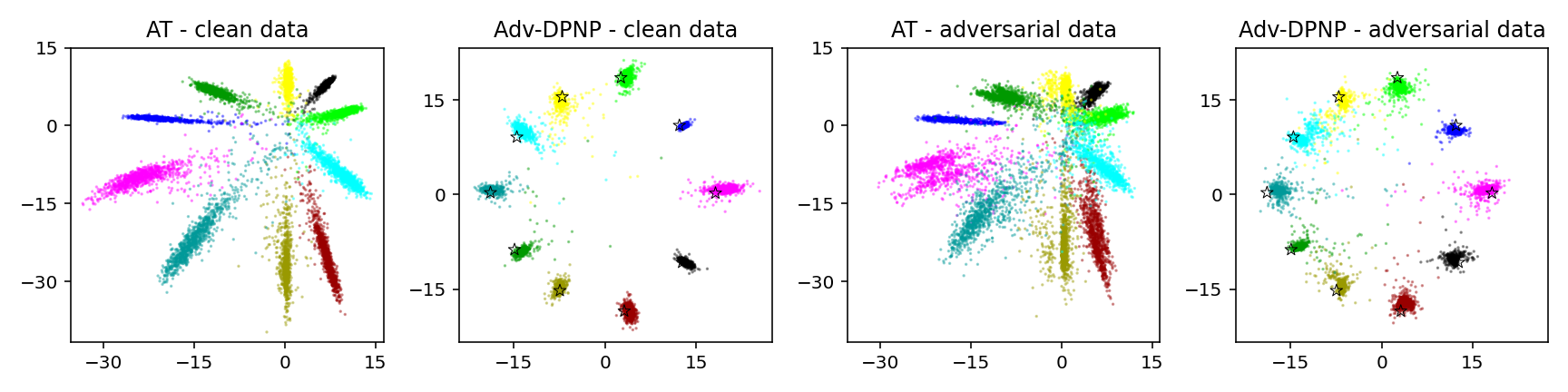}
\caption{Visualization of feature representations of LeNets++ for clean data (left two plots) and adversarial data (right two plots) using AT and Adv-DPNP. The stars indicate the learned prototypes by Adv-DPNP. The models are trained on the MNIST dataset, and the feature distributions from the test set are presented. Adversarial examples are generated using PGD with a perturbation bound of $\epsilon = 0.3$, a step size of 0.01, and 40 iterations during both training and evaluation phases.}
\label{fig:MNIST-2D}
\end{figure}

To further illustrate the discrimination in the latent space, we conduct an additional experiment on the MNIST dataset \cite{Mnist}. We employ the LeNets++ architecture \cite{Wen2016}, utilizing its two-dimensional feature representation for direct visualization.
Specifically, we train AT as a baseline model and our proposed Adv-DPNP. The resulting two-dimensional distributions of clean and adversarial features from the test dataset are visualized in Figure~\ref{fig:MNIST-2D}. 
Compared to standard AT, Adv-DPNP significantly reduces both angular and radial variations of feature representations under clean and adversarial conditions. Also, Adv-DPNP creates more distinct inter-class margins, yielding improved separation between class prototypes.
Moreover, the adversarially perturbed features in Adv-DPNP closely align with the prototypes learned from clean data, demonstrating its capability to robustly preserve the geometric structure of the feature space.

\subsection{Robustness against Common Corruptions}\label{subsec44}

\begin{figure}[t]
\centering
\includegraphics[width=1.0\textwidth]{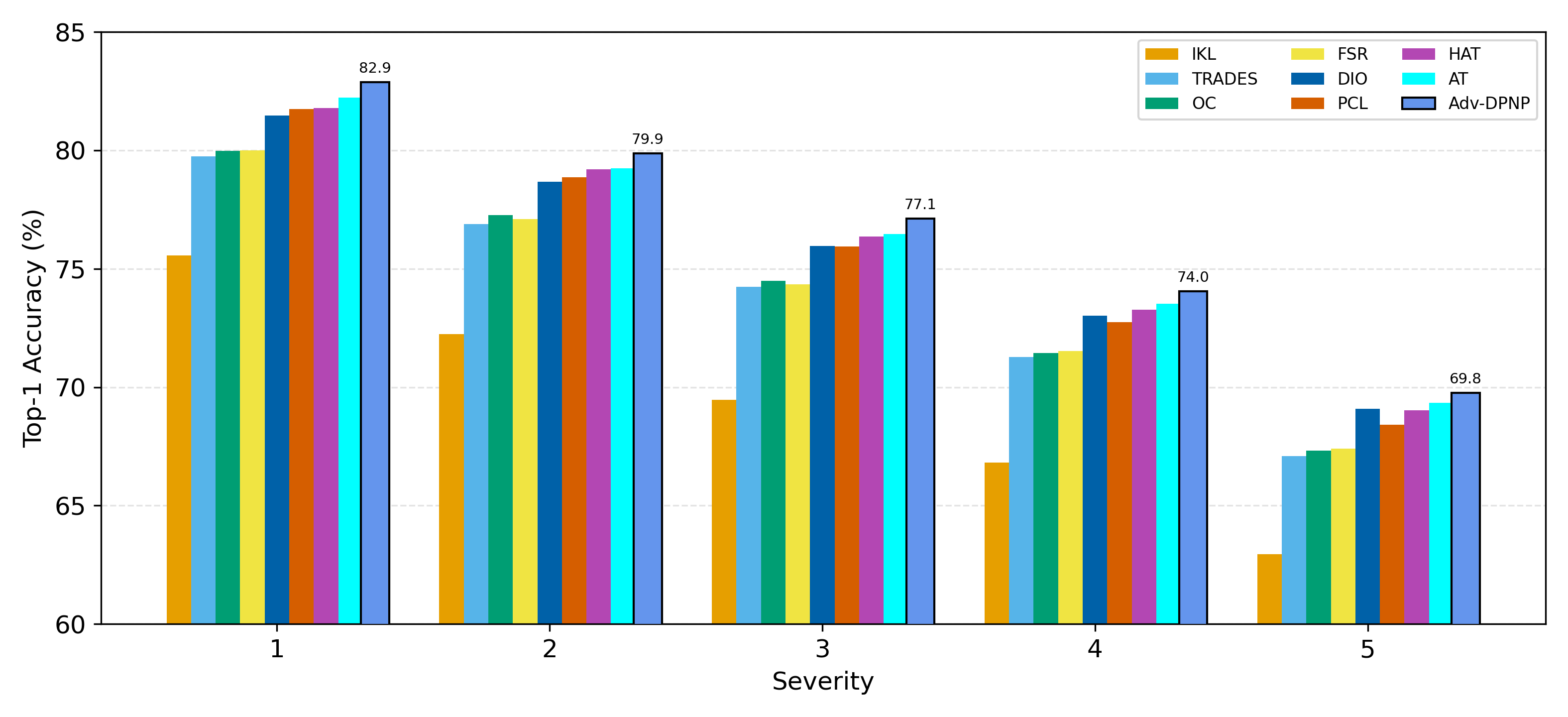}
\caption{Comparison of models' performance against five different corruption severity levels for the CIFAR-10-C dataset. Each bar represents an average of 19 corruption types for a given level of corruption strength. The Y-axis is limited to 60–85\% for readability.}
\label{fig:cifar10c}
\end{figure}

\begin{table}[b]
\centering
\scriptsize 
\setlength{\tabcolsep}{3pt}
\begin{tabular*}{\textwidth}{@{\extracolsep{\fill}} lcccccccccccc}
    \toprule
    Method & AT & AT\_HE & TRADES & MART & PCL & OC & HAT & FSR & DIO & IKL & Adv-DPNP \\
    \midrule
    Avg Acc (\%) & 76.16 & 73.20 & 73.84 & 69.04 & 75.54 & 74.09 & 75.92 & 74.06 & 75.64 & 69.40 & \textbf{76.73} \\
    \bottomrule
\end{tabular*}
\caption{Average accuracy on CIFAR-10-C. The models are adversarially trained on CIFAR-10. `Avg Acc` is the mean accuracy across all corruptions and severities.}
\label{tab:cifar10c}
\end{table}

To assess the natural robustness of our method, we evaluate the performance of adversarially trained models on the original CIFAR-10 dataset using common corruptions dataset CIFAR-10-C for test \cite{hendrycks2019robustness}. CIFAR-10-C is created by corrupting the original CIFAR-10 test images with 19 corruptions. These corruptions are categorized into four types including noise, blur, weather, and digital, with each type having five levels of severity or intensity, which increases from level one to five.

We evaluate the accuracy of the adversarially trained models from Section \ref{subsec41} on the CIFAR-10-C dataset without any additional fine-tuning. Figure \ref{fig:cifar10c} illustrates the accuracy for each level of severity averaged across all corruptions. Adv-DPNP consistently improves accuracy at each severity level and outperforms other methods. In addition, following \cite{hendrycks2019robustness}, we report the average accuracy across all severities in Table \ref{tab:cifar10c}. Adv-DPNP increases the average accuracy by 0.57\%, with AT in second place. These results demonstrate that Adv-DPNP can effectively improve model performance under common real-world corruptions, making it an effective approach for improving robustness. 

\subsection{Verifying the Absence of Gradient Obfuscation}\label{subsec45}

\begin{figure}[t]
\centering
\includegraphics[width=0.8\textwidth]{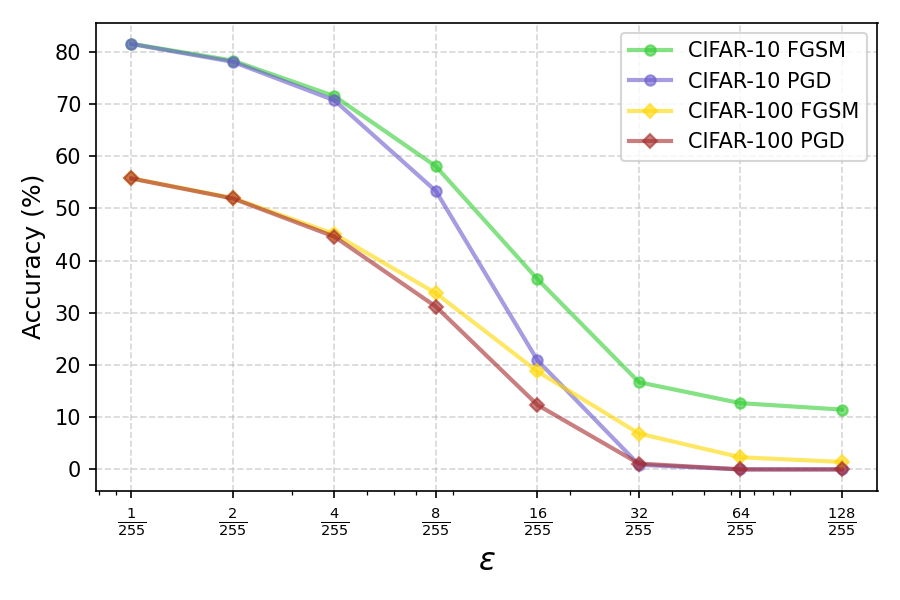}
\caption{Accuracy of the Adv-DPNP model on CIFAR-10 and CIFAR-100 under FGSM and PGD attacks across different perturbation budgets $\epsilon$. Each curve represents a specific combination of dataset and attack method (FGSM or PGD). For PGD, 20 iterations are used, and the step size is set to $\frac{2.5\,\epsilon}{20}$. As $\epsilon$ increases, accuracy decreases for all curves, with PGD ultimately exhibiting a higher success rate than FGSM.}\label{fig2}
\end{figure}

In this section, we follow the guidelines outlined in \cite{carlini2019evaluating, obfuscated-gradients} to verify the absence of gradient obfuscation in our Adv-DPNP model. For this purpose, we evaluate Adv-DPNP with the ResNet-18, trained according to Section~\ref{subsec41}, by performing a series of tests under different conditions.

\noindent \textbf{Increasing Perturbation Size:} Figure~\ref{fig2} illustrates the performance of the Adv-DPNP model on CIFAR-10 and CIFAR-100 under PGD and FGSM attacks with different perturbation sizes $\epsilon$. 
Our observations indicate that increasing the perturbation budget strictly decreases the model accuracy. In particular, PGD as an iterative attack eventually achieves a 100\% success rate at larger $\epsilon$ values, whereas the single-step FGSM attack is relatively weaker. These findings confirm that the accuracy drop is strictly tied to the perturbation size.

\begin{table}[t]
\centering
\begin{minipage}{0.5\textwidth}
\centering
\scriptsize
\begin{tabular*}{0.975\textwidth}{@{\extracolsep{\fill}}lcccc}
\toprule
\multirow{2}{*}{Iterations} & \multicolumn{2}{c}{CIFAR-10} & \multicolumn{2}{c}{CIFAR-100} \\
\cmidrule(r){2-5}
 & Acc (\%) & Loss & Acc (\%) & Loss \\
\midrule
20  & 53.33 & 1.2380 & 30.80 & 2.9387 \\
40  & 52.64 & 1.2543 & 30.56 & 2.9595 \\
100 & 52.53 & 1.2571 & 30.37 & 2.9634 \\
250 & 52.48 & 1.2575 & 30.25 & 2.9641 \\
500 & 52.45 & 1.2577 & 30.24 & 2.9643 \\
\bottomrule
\end{tabular*}
\caption{Effect of increasing the number of PGD iterations on the accuracy (\%) and CE loss of the Adv-DPNP model. The model is evaluated on CIFAR-10 and CIFAR-100 under a PGD attack configured with a perturbation bound of $\epsilon = 8/255$ and a step size set to 1/255.}
\label{tab10}
\end{minipage}
\hfill
\begin{minipage}{0.455\textwidth}
\centering
\scriptsize
\begin{tabular*}{0.95\textwidth}{@{\extracolsep{\fill}}lcc}
\toprule
\multirow{2}{*}{Random Starts} & \multicolumn{1}{c}{CIFAR-10} & \multicolumn{1}{c}{CIFAR-100} \\
\cmidrule(r){2-3}
 & Acc (\%) & Acc (\%) \\
\midrule
1  & 53.25 & 30.76 \\
5  & 52.98 & 30.45 \\
10 & 52.86 & 30.38 \\
20 & 52.79 & 30.31 \\
\bottomrule
\end{tabular*}
\caption{Effect of increasing the number of random starts on the Adv-DPNP model's accuracy (\%). The evaluation is conducted on CIFAR-10 and CIFAR-100 under a PGD attack configured with 40 iterations, a perturbation bound of $\epsilon = 8/255$, and a step size of 0.5/255.}
\label{tab11}
\end{minipage}
\end{table}

\noindent \textbf{Increasing Number of Iterations and Random Starts:} To further investigate the potential for gradient obfuscation, we report the accuracy and CE loss values when increasing the number of iterations and applying random starts for PGD attacks for CIFAR-10 and CIFAR-100, as shown in Tables~\ref{tab10} and~\ref{tab11}. The results indicate that both accuracy and loss values change marginally with increases in the number of iterations and random starts. This behavior confirms that the iterative PGD attack converges correctly.

\begin{wraptable}[9]{r}{0.42\textwidth}
\centering
\scriptsize
\begin{tabular*}{0.42\textwidth}
{@{\extracolsep{\fill}} lcc}
\toprule
 & CIFAR-10 & CIFAR-100 \\
\midrule
CE & 53.68 & 30.80 \\
Adv-DPNP & 53.39 & 30.74 \\
\bottomrule
\end{tabular*}
\caption{Comparison of accuracy (\%) on CIFAR-10 and CIFAR-100 under PGD attacks when adversarial examples are generated using the CE loss versus the composite Adv-DPNP loss.}\label{tab12}
\end{wraptable}
\noindent \textbf{Adaptive Attacks.} We also evaluate the robustness of the Adv-DPNP model by generating adversarial examples using our composite Adv-DPNP loss (Equation~\ref{eq_Adv_DPNP}) objectives for PGD attacks. 
Table~\ref{tab12} compares the model accuracy on CIFAR-10 and CIFAR-100 under CE and our composite loss. The results show only a negligible difference in accuracy between the two loss objectives, which confirms that the robustness of Adv-DPNP is not due to its loss surface making optimization difficult for attacks with CE \cite{carlini2019evaluating}.
In conclusion, the results of these experiments indicate that Adv-DPNP does not exhibit gradient obfuscation. Therefore, the Adv-DPNP model is truly robust and our evaluations are reliable.

\section{Conclusions}\label{sec5}
In this work, we proposed Adv-DPNP, an adversarial training framework that effectively incorporates discriminative prototype-based learning to enhance both adversarial robustness and clean data accuracy. By unifying class prototypes with classifier weights, and employing a dual-branch training procedure that decouples prototype updates from adversarial data, Adv-DPNP not only obtains robust representations which improve model performance under adversarial attacks but also better preserves clean accuracy and the original prototypes. In addition,
benefiting from discriminative terms in its composite loss function, our proposed method is able to offer an organized latent space, leading to better generalization.

We demonstrated that Adv-DPNP outperforms common baseline methods and state-of-the-art adversarial defenses on benchmark datasets such as CIFAR-10, CIFAR-100, and SVHN, achieving significant improvements in both clean accuracy and robustness against a wide range of adversarial perturbations. Additionally, we showed that Adv-DPNP provides superior performance under common corruptions on the CIFAR-10-C dataset.
Through extensive experiments and analysis of feature space discrimination, we validated that Adv-DPNP maintains a well-structured feature space with high intra-class compactness and inter-class separation, even when strong adversarial perturbations are applied. Furthermore, our analysis confirmed that our proposed method does not suffer from gradient obfuscation, ensuring the stability and effectiveness of the model against various attack strategies.

In conclusion, Adv-DPNP offers a promising approach for improving the adversarial robustness of DNNs while maintaining the generalization capability of the learned representations. Future work can explore and optimize the model in other domains, including natural language processing and speech recognition.

\bibliographystyle{plainnat}  
\bibliography{Adv_DPNP_bibliography.bib}

\end{document}